\documentclass[10pt,twocolumn,letterpaper]{article}

\usepackage{wacv}
\usepackage{times}
\usepackage{epsfig}
\usepackage{graphicx}
\usepackage{amsmath}
\usepackage{amssymb}
\usepackage{booktabs}
\usepackage{multirow}
\usepackage{makecell}
\usepackage{comment}
\usepackage[ruled,vlined]{algorithm2e}
\usepackage[accsupp]{axessibility}  

\def\wacvPaperID{6} 

\wacvfinalcopy 

\ifwacvfinal
\usepackage[breaklinks=true,bookmarks=false]{hyperref}
\else
\usepackage[pagebackref=true,breaklinks=true,colorlinks,bookmarks=false]{hyperref}
\fi

\begin{document}

\title{Test-time Adaptation vs. Training-time Generalization:\newline A Case Study in Human Instance Segmentation using Keypoints Estimation}


\author{Kambiz Azarian \and Debasmit Das \and Hyojin Park \and Fatih Porikli \and
Qualcomm AI Research\thanks{\noindent Qualcomm AI Research is an initiative of Qualcomm Technologies, Inc.}\\
{\tt\small \{kambiza, debadas, hyojinp, fporikli\}@qti.qualcomm.com}}

\maketitle
\thispagestyle{empty}

\begin{abstract}
    We consider the problem of improving the human instance segmentation mask quality for a given test image using keypoints estimation. We compare two alternative approaches. The first approach is a test-time adaptation (TTA) method, where we allow test-time modification of the segmentation network's weights using a single unlabeled test image. In this approach, we do not assume test-time access to the labeled source dataset. More specifically, our TTA method consists of using the keypoint estimates as pseudo labels and backpropagating them to adjust the backbone weights. The second approach is a training-time generalization (TTG) method, where we permit offline access to the labeled source dataset but not the test-time modification of weights. Furthermore, we do not assume the availability of any images from or knowledge about the target domain. Our TTG method consists of augmenting the backbone features with those generated by the keypoints head and feeding the aggregate vector to the mask head. Through a comprehensive set of ablations, we evaluate both approaches and identify several factors limiting the TTA gains. In particular, we show that in the absence of a significant domain shift, TTA may hurt and TTG show only a small gain in performance, whereas for a large domain shift, TTA gains are smaller and dependent on the heuristics used, while TTG gains are larger and robust to architectural choices.
\end{abstract}

\section{Introduction}
\begin{figure}[t]
    \begin{center}
        \includegraphics[width=0.95\linewidth]{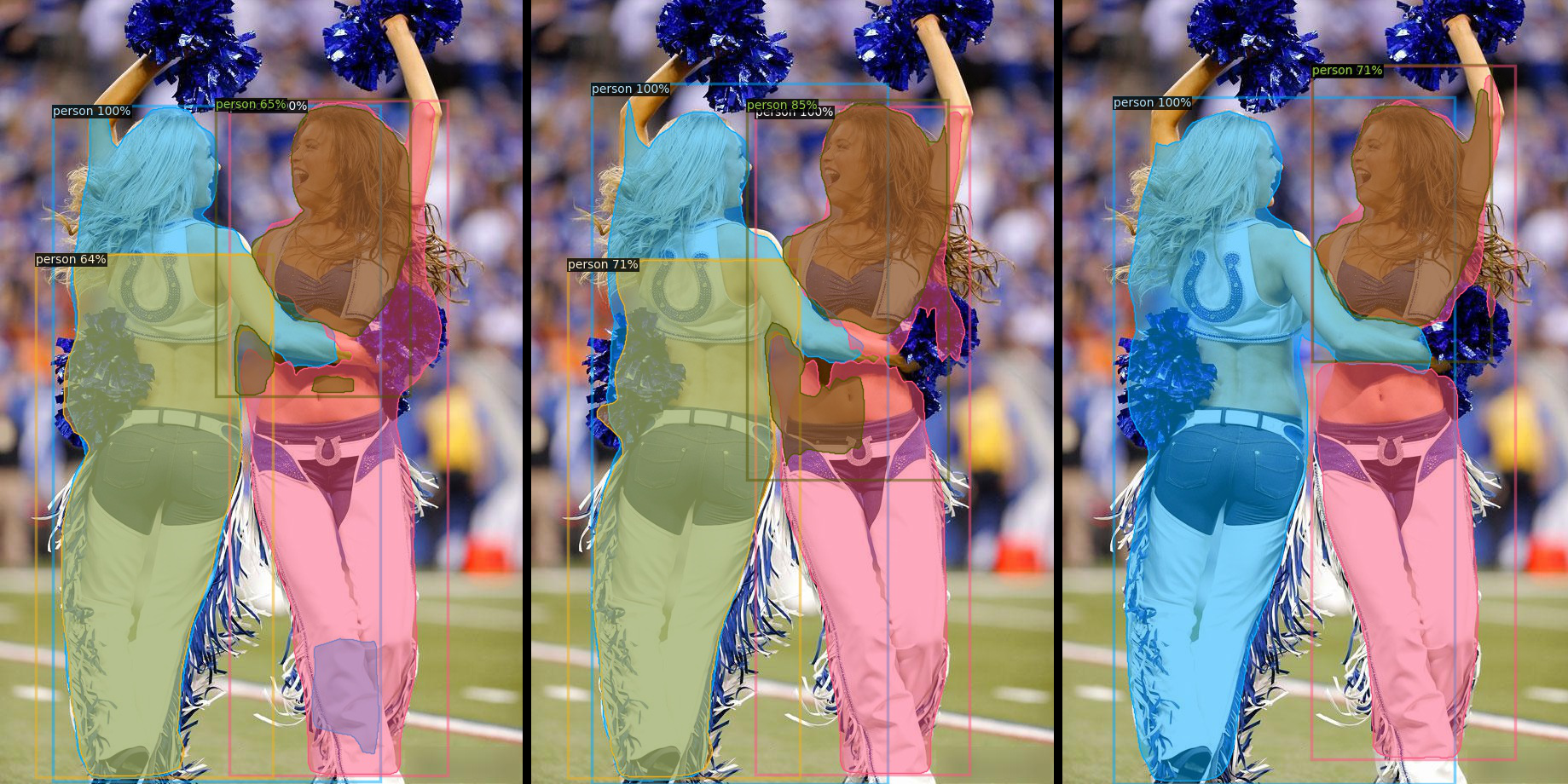}
    \end{center}
    \caption{From left to right: baseline, test-time adapted and training-time generalized Mask-RCNN.}
    \label{fig:tta_ttg_comparison}
\end{figure}

Human instance segmentation is an important task that requires finding the image regions of different individuals in a scene. This task has multiple applications ranging from video conferencing~\cite{jiang2021peopleseg,chu2022pp}, matting~\cite{ye2019edge,zhu2017fast,li2020natural}, generation~\cite{qiao2019learn,mao2021learning}, autonomous driving~\cite{zhang2016instance}, robotics~\cite{uckermann2013realtime}, extended reality~\cite{zhao2019real}, pedestrian tracking~\cite{ouyang2017jointly,mao2017can,zhou2018deep}, etc. For example, in extended reality applications, human instance segmentation can be used to detect human contours. Along with 3D information, the human contours can be used to render virtual objects at arbitrary locations, which can then enhance augmented reality experiences. For human-robot interaction, segmentation of individuals is required for understanding spatial relationships that allows the robot to plan movements and perform complex tasks. Since human instance segmentation has multiple use cases, there has been considerable effort in recent years to improve the performance of such models. 

Most methods for human instance segmentation are built on top of general instance segmentation frameworks. Such methods can be broadly divided into single-stage and multi-stage methods. Single-stage methods~\cite{dai2016instance,wang2020solo,bolya2019yolact} normally use parallel branches for detection and segmentation. The detection branch is used to localize each individual instance, while the segmentation branch learns to annotate each pixel based on the feature information densely. The output of the detection branch and the segmentation branch are employed to obtain masks for all the instances in the scene. Since detection and segmentation operate independently of each other, single-stage methods are considerably faster than multi-stage counterparts. However, the segmentation step does not utilize the localization information; thus, single-stage methods perform poorer than multi-stage ones. 

Two-stage methods~\cite{fang2021instances,liu2018path,he2017mask} generally follow a sequential approach of firstly detection and then segmentation. Multi-stage methods~\cite{chen2019hybrid,cheng2022masked} go a step further and repeat this two-step method multiple times. The detection stage normally crops out a region of interest that localizes a particular object. Subsequently, segmentation is applied for different regions of interest to produce masks for different instances in the scene. Since segmentation and detection depend on each other during training or inference, segmentation quality is much better compared to single-stage methods. Hence, in this paper, we opt for the popular two-stage framework Mask-RCNN~\cite{he2017mask}, which also carries a pose estimation head to aid human instance segmentation.

Compared to general instance segmentation, human instance segmentation is relatively more challenging for a couple of reasons. Firstly, there is large intra-class variation within the human category due to various outfits, poses, and deformability of the human body. Secondly, the number of humans and their locations in a scene can be random, which increases the complexity of model inference. Finally, humans tend to interact with each other and other objects causing occlusions and obstructions, resulting in unnatural 2D shapes and poses. To tackle such challenges, pose estimation~\cite{chen2018cascaded,fang2017rmpe,kocabas2018multiposenet} has been used to enhance human instance segmentation.

There have been different approaches to using pose estimation for human segmentation. The most popular method~\cite{zhang2019pose2seg} follows a sequential approach, where humans are firstly detected, and their corresponding poses are estimated. The sparse keypoints are then grouped, and the generated heatmaps are used to segment the humans. Although the proposed sequential approach is seminal, it has the weakness of propagating the pose estimation errors to the segmentation stage, which can negatively affect the performance, especially for occluded humans. Alternatively, there are methods~\cite{papandreou2018personlab,he2017mask} that carry out a joint estimation of pose and instance segmentation, which do not suffer from propagation issues. We build on the latter framework, i.e., Mask-RCNN~\cite{he2017mask} in this paper. Specifically, we consider how human pose estimation can be used to enhance the performance of a human instance-segmentation model (trained on a source dataset) on test images from a target dataset. Our evaluations are done on the OCHuman~\cite{zhang2019pose2seg} and COCOPersons~\cite{lin2014microsoft,zhang2019pose2seg} datasets from which we draw conclusions about the feasibility and extent of such improvement using generalization, adaptation, or both.

Our main contributions are summarized as follows:
\begin{itemize}

    \item We propose a \emph{test-time} method for adapting a human segmentation network to a single unlabelled test image. It involves backpropagating keypoint pseudo-labels to adjust the backbone weights. We devise three keypoints head variants in addition to the Mask-RCNN's keypoints head. Two of the variants are transformer based, and all augment the pose estimates to include keypoint visibility/occlusion indicators.

    \item We propose a \emph{training-time} method for enhancing the performance of a human segmentation network on new domains. It involves augmenting the backbone features with those generated by the keypoints head and using the aggregate feature vector for segmentation. We show our method achieves competitive performance despite its simplicity.

    \item We evaluate the performance of both methods and, through ablations, identify factors that limit test-time adaptation (TTA) gains. We show that for a small domain shift, TTA may hurt performance, and training-time generalization (TTG) delivers only a small gain. For a large domain shift, TTA gains are small and dependent on the heuristics used, while TTG gains are more prominent and relatively insensitive to architectural details.

\end{itemize}

\section{Related Work}\label{sec:backgroundrelated}
\noindent \textbf{Non-Adaptive Instance Segmentation:} Non-adaptive Instance segmentation methods generally have enhanced architectures either through single-stage approaches~\cite{dai2016instance,wang2020solo,bolya2019yolact} or multi-stage approaches~\cite{fang2021instances,liu2018path,he2017mask}. Single-stage methods have a distributed approach where they produce feature maps for the whole image and then extract the feature maps for each instance to produce the corresponding masks. For example, InstanceFCN~\cite{dai2016instance} generates instance-specific scoring maps and outputs instance masks using an assembly module that contains operations like repooling and mask-voting. In similar spirit, CondInst~\cite{condinst} dynamically generated convolutions conditioned on each instance to produce instance-specific segmentation masks. On the other hand, YOLACT~\cite{bolya2019yolact} is an efficient method that has parallel branches for generating fixed number of prototype masks and mask coefficients. Multi-stage methods generally have a two step procedure where firstly objects are detected using bounding boxes. Then, features are extracted from the region of interest to produce the desired masks. Mask R-CNN~\cite{he2017mask} is a popular two-stage instance segmentation framework that has an additional head for predicting segmentation masks for each region extracted from the bounding box. In this paper, we focus specifically on the Mask R-CNN architecture. QueryInst~\cite{fang2021instances} is also a recent query-based multi-stage framework that uses sequences of dynamic convolution and multi-head self-attention blocks to produce more refined instance segmentation masks. More comprehensive review of non-adaptive instance segmentation methods can be found in~\cite{gu2022review}. 

\noindent \textbf{Adaptive Instance Segmentation:} There have been very few works on adaptive instance segmentation. Most of these works target segmentation of biomedical entities like nuclei, cells etc. For example, in~\cite{liu2020unsupervised}, the authors propose a multi-step procedure to tackle domain shift that includes inpainting of images, producing domain-invariant features and a task reweighting scheme to remove source bias. In~\cite{liu2020pdam}, the authors extended the framework by adding a mechanism of feature similarity maximization. Li et al.~\cite{li2022domain} also proposed domain adaptation for nuclei segmentation by category-specific feature alignment and self-training using pseudo-labels. In addition to augmented pseudo-labelling, Hsu et al.~\cite{hsu2021darcnn} proposed to use a domain separation module as well as a self-supervised consistency loss. Recently, in~\cite{li2022domain_mitotic}, the authors propose domain adaptation for mitotic cells by aligning pixel-level feature distributions and also additional supervision through a semantic head. For adaptive human instance segmentation ~\cite{srivastav2022unsupervised} is the only work that is known to us. In this work, the authors propose to jointly estimate pose and instance masks of clinicians by adapting from a source dataset. Adaptation is carried out by a feature normalization strategy and self-training procedure where pseudo-labels are refined using geometric consistency of augmentations. 

\noindent \textbf{Human Instance Segmentation:} Human instance segmentation involves segmenting humans with the aid of additional information like pose. Earlier works that used pose to segment out human instances include Pose2Seg~\cite{zhang2019pose2seg}, PersonLab~\cite{papandreou2018personlab} and Pose2Instance~\cite{tripathi2017pose2instance}. Pose2Seg is a two-stage framework where keypoint heatmaps are initially generated, which are then transformed and aggregated to pass through a segmentation decoder and produce human instance masks. Similarly, PersonLab detected keypoints but then grouped into masks using a geometric embedding descriptor. In Pose2Instance, the authors used the distance transform of keypoints as priors for human masks. More recent works on human instance segmentation include LSNet~\cite{zhang2021location} and PosePlusSeg~\cite{Ahmad_Khan_Kim_Lee_2022}. In LSNet, the authors use pose attention module and keypoint sensitive combination to aggregate information from multiple sampling points. In the PosePlusSeg framework, the authors proposed a refinement network for improved quality of poses and instances obtained from separate heads. In this paper, we use a more simpler approach where we have multiple heads for pose estimation and instance segmentation but use pose estimation task to select relevant features for adaptation and generalization. Additional works on human instance segmentation include ideas such as self-supervised consistency of human structures across videos~\cite{jiang2020self}, iterative refinement using pose with shape prior and part attention~\cite{zhou2020poseg}, deformable convolutions with geometric transformation between keypoint offsets~\cite{bai2019acpnet}.

\section{Method}\label{sec:method}
We first describe the task of human segmentation using keypoints, and then we detail our TTA and TTG approaches. Finally, we describe the heuristics and keypoints head variants we devised to enhance TTA and TTG gains.

\subsection{Human Segmentation using Keypoints}
\begin{figure}[t]
    \begin{center}
        \includegraphics[width=0.65\linewidth]{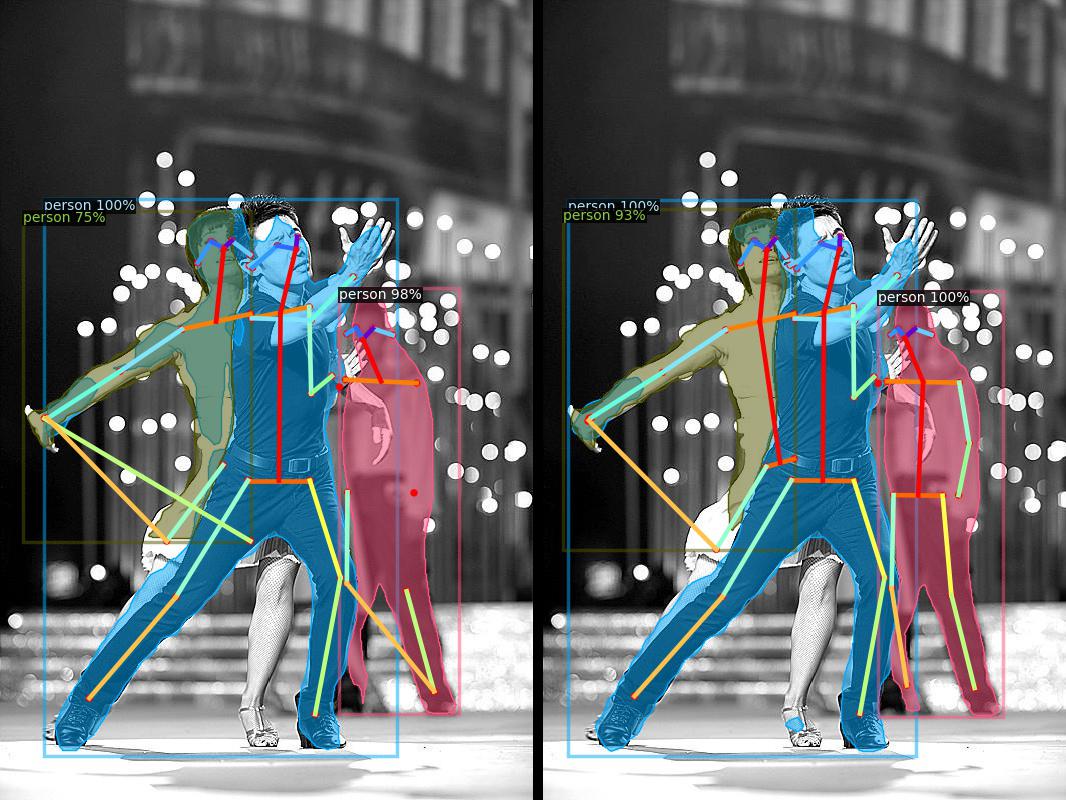}
    \end{center}
    \caption{High quality keypoints estimates can improve human segmentation masks (left to right: baseline and TTA Mask-RCNN, where the man's mask spill-over has significantly been reduced).}
    \label{fig:keypoints_help}
\end{figure}

\begin{figure*}[h]
\centering
\includegraphics[width=0.75\linewidth]{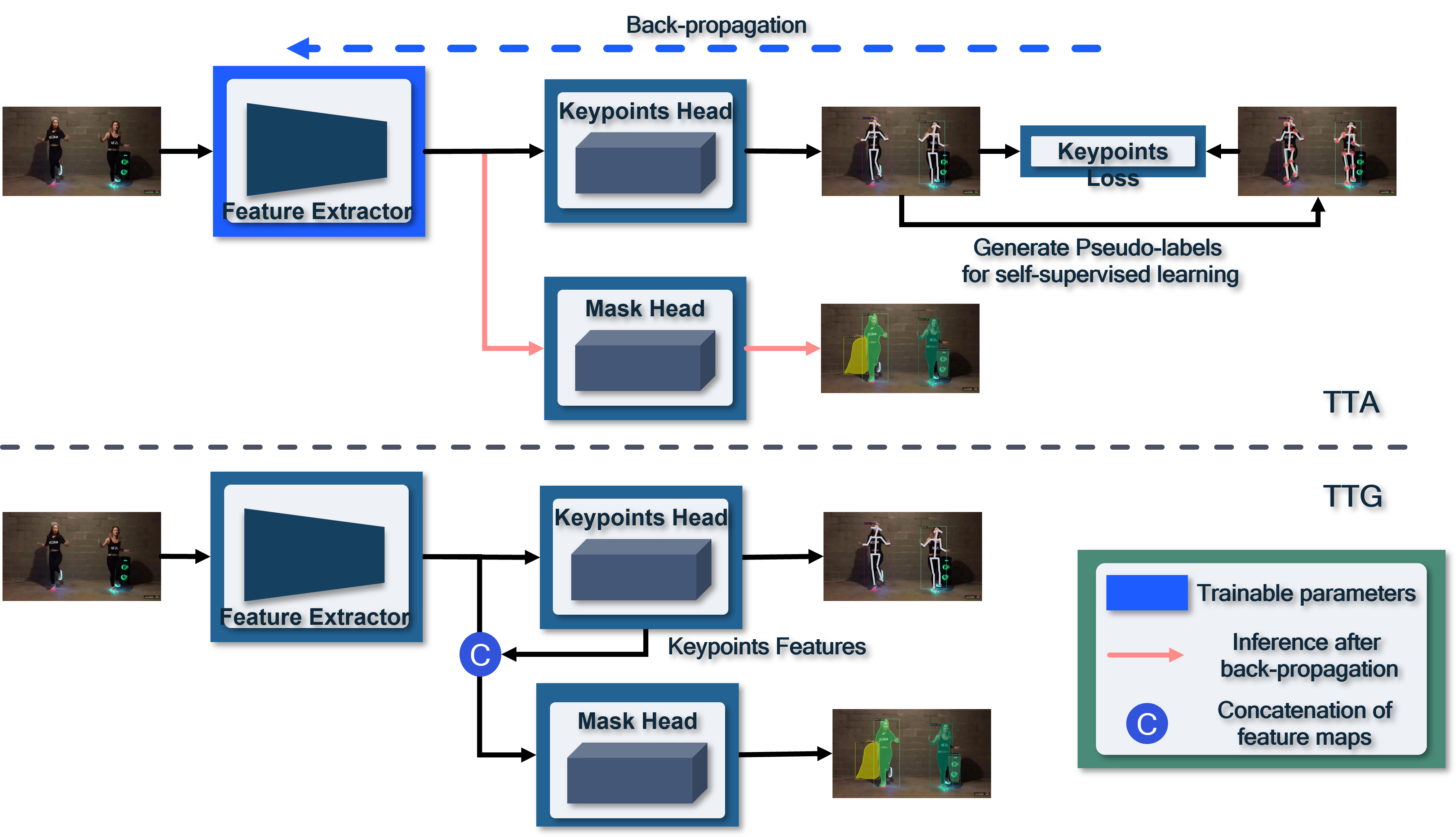}
\caption{Test-time adaptation (top) and training-time generalization (bottom) frameworks for enhancing human segmentation using pose estimates.}
\label{fig:tta_ttg_diagram}
\end{figure*}

Segmentation networks may perform poorly especially if the test image exhibits a large domain shift with respect to the training dataset, e.g., indoor vs. outdoor or daylight vs. nighttime. For \emph{human} instance segmentation, such a domain shift may be due to severe occlusion of human subjects. Figure \ref{fig:keypoints_help} depicts a case where the baseline network's mask for the man significantly spills over to that of the woman. Interestingly, despite the network's poor segmentation performance, it estimates the human keypoints fairly well. This observation motivated us to devise methods for enhancing human instance segmentation using keypoints estimates. We consider two such methods, i.e., TTA and TTG (c.f., Figure \ref{fig:tta_ttg_diagram}). Since we need pose estimates, we consider (the rather common) architecture (e.g., Mask-RCNN) where the network, in addition to its instance-segmentation head, has a separate keypoints head, with a common backbone.

\subsection{Test-time Adaptation} \label{subsec:tta}
Figure \ref{fig:tta_ttg_diagram} (top) depicts our test-time-adaptation method. In this approach we assume availability of a fully trained human instance segmentation/keypoints estimation network, i.e., $\mathbf{m} = \{ \mathbf{m}_{b}, \mathbf{m}_{m}, \mathbf{m}_{k} \}$, and a single unlabelled test image. We also allow for \emph{test-time} adaptation of network weights using backpropagation, but not access to the source dataset. These assumptions make our TTA setup very realistic, but simultaneously very challenging. 

We give the steps of our TTA method in Algorithm \ref{alg:tta}. It includes multiple rounds (e.g., $3$) of weight adaptation, each consisting of converting the keypoints estimates to pseudo-labels, plugging them in the keypoints loss together with the keypoint estimates, and backpropagating the resulting self-supervised loss to adjust the backbone weights. While our method works for any pseudo-label conversion method, $\mathbf{f}$, and keypoint loss, $\mathcal{L}_{key}$, we use simple candidates, i.e., for person bounding boxes scoring above $0.5$, we take keypoints with a minimum probability of $0.05$, and declare the location in their $56\times56$ heat-maps with the highest value as pseudo-labels. We use the multi-category cross-entropy as our keypoint loss. Finally, adapted segmentation masks are generated by running the mask head on the adapted feature map. 

\setlength{\textfloatsep}{10pt} 
\begin{algorithm}[h]
\small
\SetAlgoLined
 \textbf{Given:} Model $\mathbf{m} = \{ \mathbf{m}_{b}, \mathbf{m}_{m}, \mathbf{m}_{k} \}$, test-image $x^{tgt}$   \\
 \textbf{Step 1 (\underline{test-time}):} Adapt backbone to test image \\
 \quad \textbf{Initialize} $n$, $i \leftarrow 0$, $\mathbf{m}_{b}^0 \leftarrow \mathbf{m}_{b}$ \\
 \quad \textbf{For} $i < n$ \\
 \quad \quad \textbf{Estimate keypoints}, i.e., $y_{key}^i = \mathbf{m}_{k} \big ( \mathbf{m}_{b}^i(x^{tgt}) \big ) $ \\
 \quad \quad \textbf{Generate pseudo-labels}, i.e., $\Tilde{y}_{key}^i = \mathbf{f} \big ( y_{key}^i \big )$ \\
 \quad \quad \textbf{Update backbone}, i.e., compute $\mathbf{m}_{b}^{i+1}$ by back- \\ \quad \quad \quad propagating self-supervized $\mathcal{L}_{key}(y_{key}^i, \Tilde{y}_{key}^i)$ \\
 \quad  \textbf{Save adapted model}, i.e, $\mathbf{m}_{b}^{\text{TTA}} \leftarrow \mathbf{m}_{b}^{n}$ \\ 
 \textbf{Step 2 (\underline{test-time}):} Get TTA masks for test image \\
 \quad \textbf{Infer TTA masks}, i.e., $y_{mask}^{\text{TTA}} = \mathbf{m}_{m} \big ( \mathbf{m}_{b}^{\text{TTA}}(x^{tgt}) \big ) $ \\
\caption{TTA (test-time adaptation)}
\label{alg:tta}
\end{algorithm}

\subsection{Training-time Generalization} \label{subsec:tta}
Figure \ref{fig:tta_ttg_diagram} (bottom) shows our training-time generalization method. We assume availability of a human segmentation and pose estimation network, i.e., $\mathbf{m} = \{ \mathbf{m}_{b}, \mathbf{m}_{m}, \mathbf{m}_{k} \}$ (which does not need to be trained), and allow offline access to the labelled source dataset. We do not, however, assume the availability of any test images or knowledge about the target domain. We do not allow test-time adaptation of network weights either. These assumptions make our TTG setup both realistic and challenging. 

We give the steps of our TTG method in Algorithm \ref{alg:ttg}. Training-time generalization consists of splitting the keypoints head into two subnets, i.e., $\mathbf{m}_{k} = \mathbf{m}_{k}^{\text{reg}} \circ \mathbf{m}_{k}^{\text{fe}}$, for feature extraction and regression, respectively (the split details are given in the sequel). The mask head also needs to be modified to accommodate the extra keypoints features. The TTG model $\mathbf{m}^{\text{TTG}} = \{ \mathbf{m}_{b}, \mathbf{m}_{m}^{\text{TTG}}, \mathbf{m}_{k} \}$ is then trained on the labeled source dataset using the original segmentation and pose losses. At test-time, the segmentation masks are generated by running the TTG mask head on the aggregated feature map. 

\setlength{\textfloatsep}{10pt} 
\begin{algorithm}[h]
\small
\SetAlgoLined
 \textbf{Given:} Model $\mathbf{m} = \{ \mathbf{m}_{b}, \mathbf{m}_{m}, \mathbf{m}_{k} \}$, source-dataset $\mathcal{X}^{src} = \{ (x^{src}_i, t^{src}_i) \}$ \\
\textbf{Step 1 (\underline{training-time}):} Train generalized model on source dataset \\
\quad \textbf{Split keypoints head} into feature-extractor \& regressor \\ \quad \quad subnets, i.e., $\mathbf{m}_{k} = \mathbf{m}_{k}^{\text{reg}} \circ \mathbf{m}_{k}^{\text{fe}}$ \\
\quad \textbf{Modify mask head}, i.e., $\mathbf{m}_{m}^{\text{TTG}}$ to accommodate the \\ \quad \quad extra keypoints features, i.e., $\mathbf{m}_{k}^{\text{fe}} \circ \mathbf{m}_{b}(x^{tgt})$ \\
\quad \textbf{Train TTG model}, i.e., $\mathbf{m}^{\text{TTG}} = \{ \mathbf{m}_{b}, \mathbf{m}_{m}^{\text{TTG}}, \mathbf{m}_{k} \}$ on \\ \quad \quad  $\mathcal{X}^{src} = \{ (x^{src}_i, t^{src}_i) \}$ \\
\textbf{Step 2 (\underline{test-time}):} Get TTG masks for test image $x^{tgt}$ \\
 \quad \textbf{Infer TTG masks} using aggregated features, i.e., \\ \quad \quad $y_{mask}^{\text{TTG}} =$ $\mathbf{m}_{m}^{\text{TTG}} \big ( \mathbf{m}_{b}(x^{tgt}), \mathbf{m}_{k}^{\text{fe}} \circ \mathbf{m}_{b}(x^{tgt}) \big ) $ \\
\caption{TTG (training-time generalization)}
\label{alg:ttg}
\end{algorithm}

\subsection{Heuristics and Keypoints Head Variants} \label{subsec:variants}
The performance of the TTA and TTG methods greatly depends on the quality of the pseudo-labels and features generated by the keypoints head, respectively. This motivates us to devise three keypoints head variants besides the Mask-RCNN's original. In the following, we discuss each variant and the heuristics behind it. Note that the input feature map from the backbone to all these heads is of dimension $N \times 256 \times 14 \times 14$, where $N$ is the number of person bounding boxes.

\begin{figure}[t]
    \begin{center}
        \includegraphics[width=0.65\linewidth]{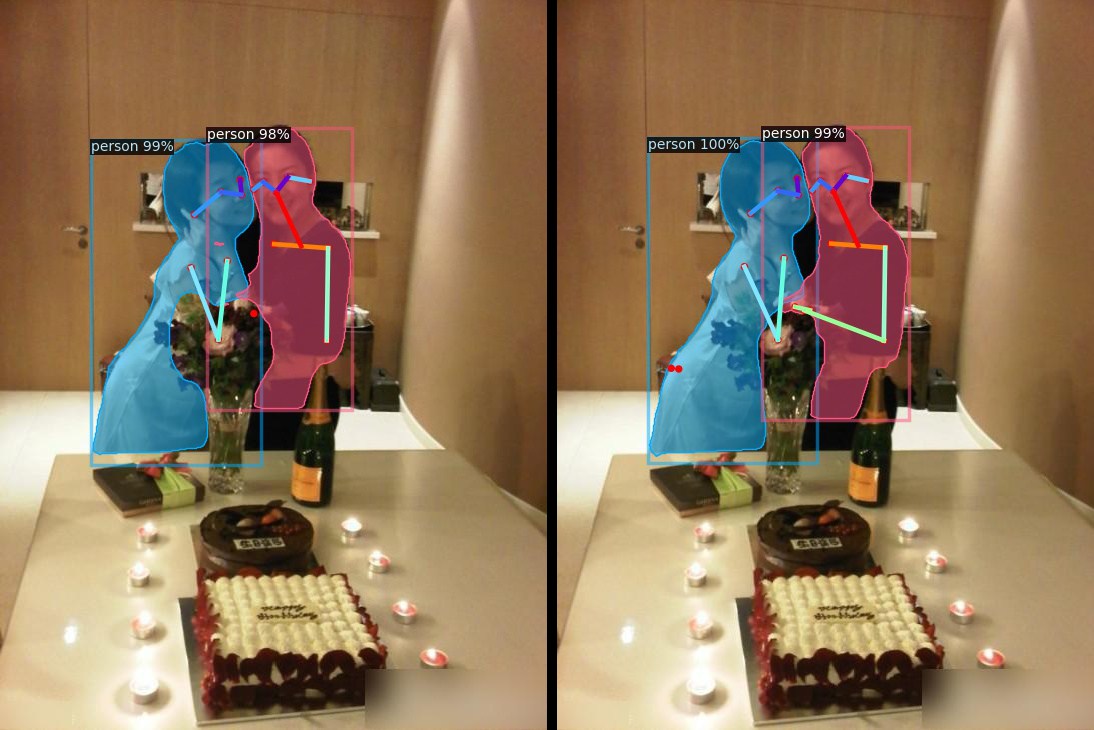}
    \end{center}
    \caption{Backpropagating the left person's elbow pseudo-label, without specifying it as occluded, degrades TTA performance by inclusion of part of the bouquet as that person's mask (left to right: baseline and TTA for Mask-RCNN).}
    \label{fig:visibility_is_necessary}
\end{figure}

\begin{figure}[t]
    \begin{center}
        \includegraphics[width=0.65\linewidth]{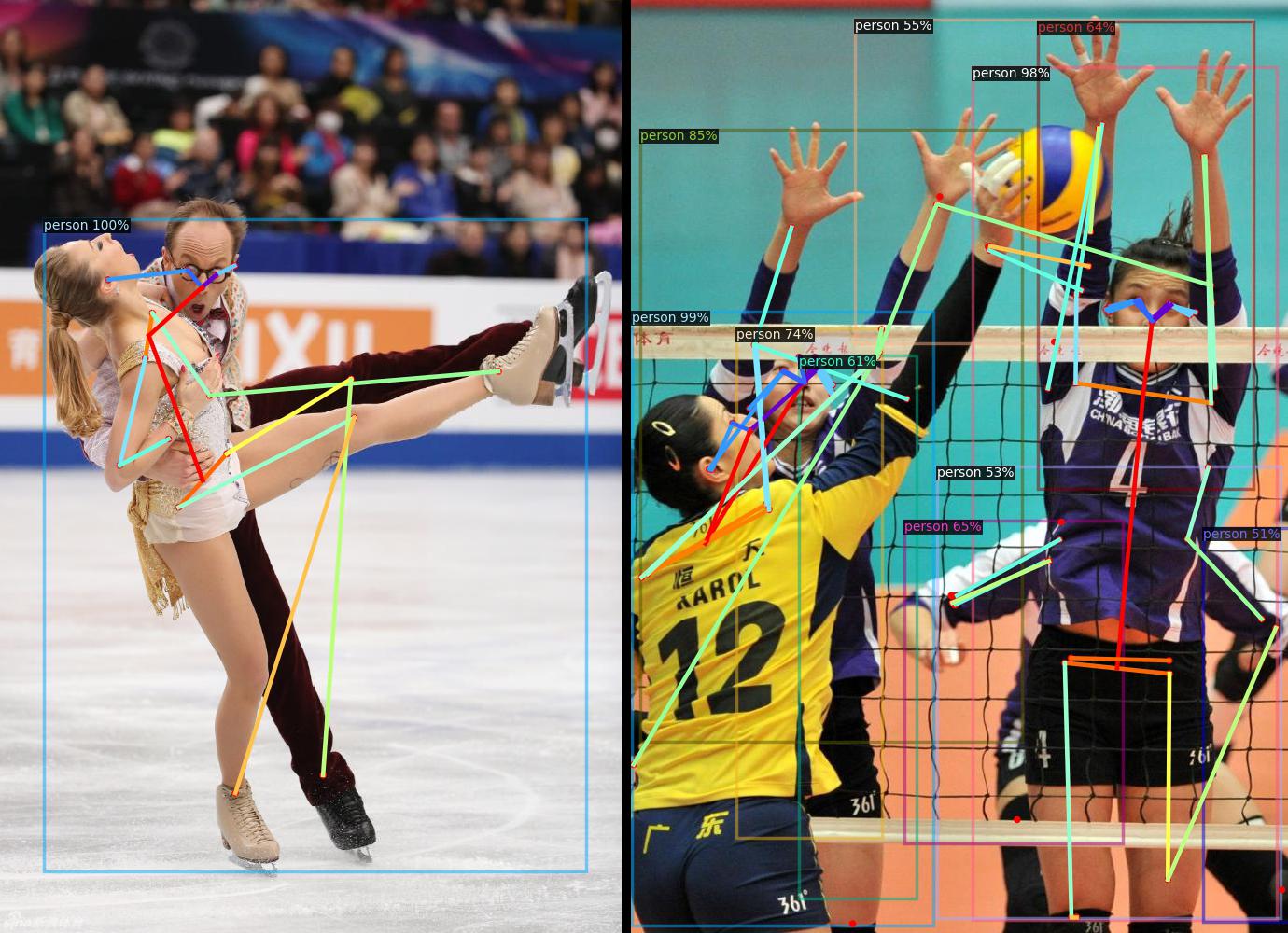}
    \end{center}
    \caption{Keypoint estimates from a convolutional head can be of very low quality.}
    \label{fig:messed_up_keypoints}
\end{figure}

\begin{itemize}
    \item \textbf{Mask-RCNN:} $\mathbf{m}_{k}^{\text{fe}}$ consists of $8$ 2d-convolutional layers, each with $512$ output channels, a $3\times3$ kernel size and a stride of $1$. Hence, there are $512$ extra keypoints features. $\mathbf{m}_{k}^{\text{reg}}$ consists of a 2d-transposed-convolutional layer with $17$ (i.e., number of keypoints) output channels, a $4\times4$ kernel and a stride of $2$, followed by a bilinear upsampler to increase the keypoints heatmap resolution to $N \times 17 \times 56\times56$.
    
    \item \textbf{Variant1:} While the Mask-RCNN keypoints head estimates the keypoint positions, it does not predict whether they are visible or occluded. This may hurt the TTA performance, e.g., in Figure \ref{fig:visibility_is_necessary}, the position of the left person's elbow has correctly been estimated, however, not specifying it as occluded has caused parts of the bouquet to be included in the TTA mask. To address this all variants predict if a keypoint is visible or occluded, e.g., Variant 2 is identical to Mask-RCNN except that the 2D-transposed-convolutional layer has $51(=3\times17)$ outputs to allow two additional outputs per location for visible/occluded prediction.

    \item \textbf{Variant2:} As Figure \ref{fig:messed_up_keypoints} shows, the keypoint estimates from a convolutional head can be of very low quality, severely impacting the TTA/TTG gains. To enhance keypoint estimation (through global attention) Variant2 uses transformers, i.e., $\mathbf{m}_{k}^{\text{fe}}$ consists of a transformer decoder with $6$ layers, $8$ heads and $17$ queries of width $256$, operating on the backbone feature map. The input queries are trainable parameters, while the output queries are mapped to keypoints and decoded by a 3-layer MLP to $N \times 51 \times 14 \times 14$ keypoints heatmap (after reshaping). Hence, there are $51$ extra keypoints features. $\mathbf{m}_{k}^{\text{reg}}$ consists of a bilinear upsampler to increase the heatmap resolution to $56 \times 56$.

    \item \textbf{Variant 3:} As will be shown (c.f., Table \ref{tab_keypoints}), while Variant 2 shows slightly better (i.e., $1\%$) $\textbf{AP}_{key}$ numbers on the target datasets, it lags Mask-RCNN and Variant 1 on the source dataset by a large margin ($4\%$). This is because in Variant 2, the transformer output queries are directly decoded to $14 \times 14$ heatmaps, without using convolutional layers. Variant 3 gets around this problem by using the last transformer layer's value projections, and attention weights to form a separate $15 \times 14 \times 14$ feature map for each keypoint. Hence there are $(255 = 17 \times 15)$ extra keypoints features. $\mathbf{m}_{k}^{\text{reg}}$ consists of two group 2D-convolutional and one group 2D-transposed-convolutional layers that further process keypoint feature maps, independently from one another, followed by a bilinear upsampler to get the $56 \times 56$ heatmaps.    
\end{itemize}

\section{Experiments}\label{sec:experiments}
This section consists of experimental details, TTA vs. TTG comparison results and ablation studies.

\subsection{Experimental Details}
We report evaluation results on the COCOPersons~\cite{zhang2019pose2seg} and OCHuman~\cite{zhang2019pose2seg} datasets. The COCOPersons dataset, which consists of 60K images, is a refined split obtained from MS-COCO~\cite{lin2014microsoft}. In this split, non-person categories as well as person categories with small annotations are removed since they do not contain keypoint annotations. The OCHuman dataset is highly challenging and mainly consists of occluded humans, e.g., the average MaxIOU for each person in COCOPersons is 0.08 while that of OCHuman is 0.67. Both these datasets contain 17 keypoints. 

We use Mask-RCNN \cite{he2017mask} for our TTA and TTG experiments, though our methods apply to any architecture consistent with Figure \ref{fig:tta_ttg_diagram}. More specifically, we use the detectron2 \cite{wu2019detectron2} codebase with ResNet-50-FPN and ResNet-101-FPN as backbones and the mask and keypoints losses implemented therein. The three variants in this work differ from the standard Mask-RCNN only in their keypoints head as detailed in Section \ref{sec:method}.

We train all models and variants offline on the COCOPersons \emph{train} set. For baseline and TTG experiments, we keep model weights frozen throughout evaluations on COCOPersons \emph{val} split and OCHuman \emph{val} and \emph{test} splits. For TTA experiments, we reset the weights to their pre-adaptation state before evaluating each test image. We then adapt the model to the image $n=3$ times with a TTA learning rate of $1\mathrm{e}{-3}$, after which we segment the image to get the TTA person masks (c.f. Algorithm \ref{alg:tta}). This process is repeated for each test image in the dataset. We report $\textbf{AP}_{key}$ and $\textbf{AP}_{mask}$ mean and standard deviation across multiple runs ($4$ and $3$ seeds for ResNet50-FPN and ResNet-101-FPN, respectively).

\subsection{TTA vs. TTG Results}
\begin{figure}[t]
    \begin{center}
        \includegraphics[width=0.95\linewidth]{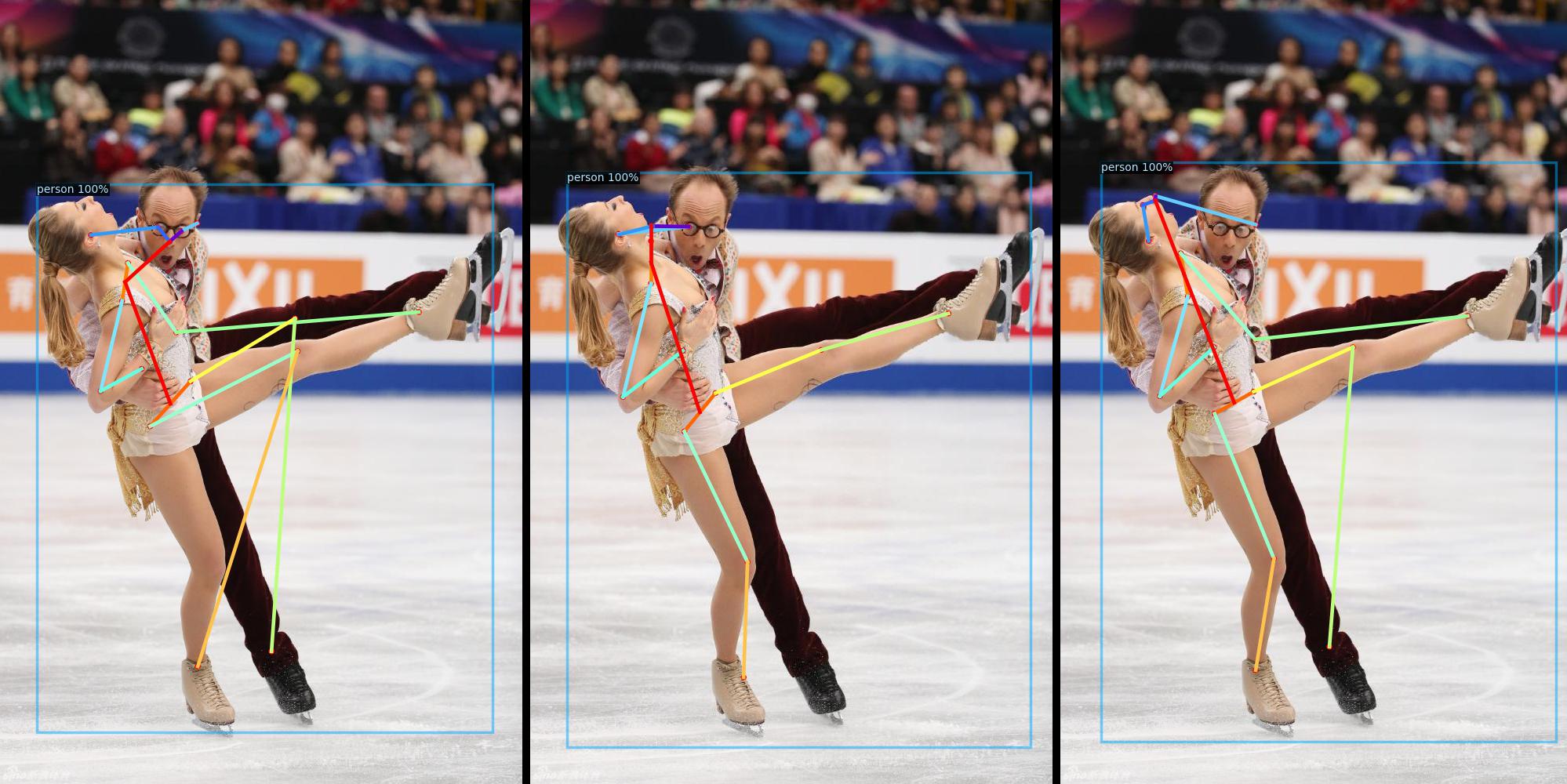}
    \end{center}
    \caption{Keypoints pseudo-label quality is a major limiting factor for TTA. From left to right: Variant 1, 2 and 3.}
    \label{fig:keypoints_heads_comparison}
\end{figure}

Table \ref{tab_keypoints} gives $\textbf{AP}_{key}$ numbers for Mask-RCNN and its variants. The large drop, i.e., more than $30\%$, in Mask-RCNN's $\textbf{AP}_{key}$ when moving from COCOPersons \emph{val} to OCHuman \emph{val} and \emph{test} attests to the significant domain shift between these datasets. Variant $1$ uses the same keypoints head as Mask-RCNN, hence its numbers are similar. For the source dataset, i.e., COCOPersons \emph{val}, Variant $2$ is lagging Mask-RCNN and Variant $3$'s $\textbf{AP}_{key}$ by $4\%$ to $\%5$, however it shows around $1\%$ advantage over the target datasets, i.e., OCHuman \emph{val} and \emph{test}. This is likely due to Variant $2$ not using convolutional layers and directly decoding keypoint tokens into spatial heatmaps. Figure \ref{fig:keypoints_heads_comparison} depicts the keypoints estimates from the three variants for a test image. As both Table \ref{tab_keypoints} and Figure \ref{fig:keypoints_heads_comparison} suggest, use of transformers does not improve the quality of keypoint pseudo-labels, which is a major factor limiting TTA gains. 

\begin{table}[t]
	\centering
		\caption{\small Keypoint head's $\textbf{AP}_{key}$ (ResNet-50-FPN, 4 seeds). The large difference ($>30\%$) between COCOPersons \emph{val} and OCHuman \emph{val} and \emph{test} shows the large domain-shifts involved. Variant 2, despite a weaker performance on source generalizes better ($1\%$) to the target domains, thanks to using transformers.}
	\resizebox{0.4\textwidth}{!}{    
{
\begin{tabular}{|c|c|c|c|}
\hline
\multirow{2}{*}{\textbf{Model}}     & \multirow{2}{*}{\bfseries\makecell{COCO \\Persons \emph{val}}}     & \multirow{2}{*}{\bfseries\makecell{OCHuman\\ \emph{val}}}     & \multirow{2}{*}{\bfseries\makecell{OCHuman\\ \emph{test}}} \\
&   &   &   \\ \hline
Mask-RCNN     & \makecell{64.85\\(0.10)}  & \makecell{32.21\\(0.32)}     & \makecell{31.91\\(0.23)} \\ \hline
Variant 1     & \makecell{64.87\\(0.17)}  & \makecell{32.19\\(0.17)}     & \makecell{31.83\\(0.32)} \\ \hline
Variant 2     & \makecell{59.90\\(0.11)}  & \makecell{33.45\\(0.54)}     & \makecell{32.91\\(0.23)} \\ \hline
Variant 3     & \makecell{63.94\\(0.12)}  & \makecell{32.30\\(0.31)}     & \makecell{31.67\\(0.28)} \\ \hline
\end{tabular}}}
	\label{tab_keypoints}
\end{table}

Table \ref{tab_coco_r50} gives $\textbf{AP}_{mask}$ numbers for various models over the source dataset, i.e., COCOPersons \emph{val}. Note that while TTG improves $\textbf{AP}_{mask}$ by around $1\%$ percent, TTA degrades it by about the same amount. TTG improves $\textbf{AP}_{mask}$ by using extra, i.e., keypoint, ground-truth labels for training the \emph{segmentation} head. TTA hurts $\textbf{AP}_{mask}$ because there is not enough domain shift between COCOPersons \emph{train} and \emph{val} to justify adjusting the model's already optimized weights based on a single unlabelled test image.  

\begin{table}[t]
	\centering
		\caption{\small $\textbf{AP}_{mask}$ on COCOPersons \emph{val} (ResNet-50-FPN, 4 seeds). Without a large domain-shift TTA hurts performance and TTG shows only a small gain ($1\%$).}
	\resizebox{0.3\textwidth}{!}{    
{
\begin{tabular}{|c|c|c|c|}
\hline
\multirow{2}{*}{\textbf{Model}}     & \multirow{2}{*}{\bfseries\makecell{Baseline}}     & \multirow{2}{*}{\bfseries\makecell{\textbf{TTA}}}     & \multirow{2}{*}{\bfseries\makecell{\textbf{TTG}}} \\
&   &   &   \\ \hline
Mask-RCNN     & \makecell{60.11\\(0.12)} & \makecell{58.59\\(0.25)} & \makecell{61.25\\(0.12)} \\ \hline
Variant 1     & \makecell{60.22\\(0.25)} & \makecell{58.62\\(0.15)} & \makecell{61.45\\(0.10)} \\ \hline
Variant 2     & \makecell{59.55\\(0.04)} & \makecell{58.55\\(0.04)} & \makecell{60.80\\(0.15)} \\ \hline
Variant 3     & \makecell{60.12\\(0.15)} & \makecell{59.02\\(0.24)} & \makecell{60.97\\(0.03)} \\ \hline
\end{tabular}}}
	\label{tab_coco_r50}
\end{table}

Tables \ref{tab_ocval_r50} and \ref{tab_octest_r50} give $\textbf{AP}_{mask}$ numbers for the target datasets, i.e., OCHuman \emph{val} and \emph{test}. Here the domain shift is large enough (i.e., more than $40\%$ drop in $\textbf{AP}_{mask}$) to enable TTA to improve various variants' performances using a single unlabelled image. The largest gain is around $1\%$ and corresponds to Variant $2$, due to this variant's slightly higher $\textbf{AP}_{key}$. It is noteworthy that TTA has little, if any benefit, for Mask-RCNN as it does not distinguish between the visible and occluded keypoints as discussed earlier. More importantly though, the TTG gains are significantly higher than those offered by TTA, e.g., $3.77\%$ for TTG compared to TTA's $1.05\%$ for Variant $2$ on OCHuman \emph{val}. Furthermore, the TTG gains for all variants are similar, i.e., TTG $\textbf{AP}_{mask}$ numbers are within a standard deviation from one another. This is because TTG does not use the keypoint \emph{estimates} directly; it instead uses the keypoint \emph{features} that are richer in information, e.g., implicitly infer keypoints' visibility/occlusion. It also reconfirms that TTG gain comes from using the extra keypoint ground-truth labels for training the segmentation head and not from the details of the keypoint head architecture. 

\begin{table}[t]
	\centering
		\caption{\small $\textbf{AP}_{mask}$ on OCHuman \emph{val} (ResNet-50-FPN, 4 seeds). With a large domain-shift TTA improves all variants, however TTA gains are smaller than TTG, and more sensitive to keypoint head's architecture.}
	\resizebox{0.3\textwidth}{!}{    
{
\begin{tabular}{|c|c|c|c|}
\hline
\multirow{2}{*}{\textbf{Model}}     & \multirow{2}{*}{\bfseries\makecell{Baseline}}     & \multirow{2}{*}{\bfseries\makecell{\textbf{TTA}}}     & \multirow{2}{*}{\bfseries\makecell{\textbf{TTG}}} \\
&   &   &   \\ \hline
Mask-RCNN     & \makecell{17.74\\(0.25)} & \makecell{17.66\\(0.15)} & \makecell{21.48\\(0.10)} \\ \hline
Variant 1     & \makecell{17.93\\(0.19)} & \makecell{18.00\\(0.19)} & \makecell{21.56\\(0.17)} \\ \hline
Variant 2     & \makecell{18.15\\(0.27)} & \makecell{18.79\\(0.28)} & \makecell{21.51\\(0.51)} \\ \hline
Variant 3     & \makecell{18.41\\(0.29)} & \makecell{18.44\\(0.29)} & \makecell{21.20\\(0.14)} \\ \hline
\end{tabular}}}
	\label{tab_ocval_r50}
\end{table}

\begin{table}[t]
	\centering
		\caption{\small $\textbf{AP}_{mask}$ on OCHuman \emph{test} (ResNet-50-FPN, 4 seeds). Same trends as Table \ref{tab_ocval_r50}, i.e., TTA gains are smaller and more heuristic dependent than TTG.}
	\resizebox{0.3\textwidth}{!}{    
{
\begin{tabular}{|c|c|c|c|}
\hline
\multirow{2}{*}{\textbf{Model}}     & \multirow{2}{*}{\bfseries\makecell{Baseline}}     & \multirow{2}{*}{\bfseries\makecell{\textbf{TTA}}}     & \multirow{2}{*}{\bfseries\makecell{\textbf{TTG}}} \\
&   &   &   \\ \hline
Mask-RCNN     & \makecell{17.48\\(0.15)} & \makecell{17.51\\(0.15)} & \makecell{21.30\\(0.18)} \\ \hline
Variant 1     & \makecell{17.42\\(0.33)} & \makecell{17.69\\(0.22)} & \makecell{21.09\\(0.07)} \\ \hline
Variant 2     & \makecell{17.92\\(0.21)} & \makecell{18.50\\(0.11)} & \makecell{20.93\\(0.08)} \\ \hline
Variant 3     & \makecell{17.99\\(0.06)} & \makecell{18.04\\(0.12)} & \makecell{20.64\\(0.21)} \\ \hline
\end{tabular}}}
	\label{tab_octest_r50}
\end{table}

Tables \ref{tab_coco_r101}, \ref{tab_ocval_r101} and \ref{tab_octest_r101} give our results for ResNet-101-FPN backbone, and show very similar trends, i.e., in the absence of a significant domain shift, TTA may hurt and TTG show only a small gain (c.f., Table \ref{tab_coco_r101}), whereas with a large enough domain shift, TTA gains are smaller and more dependent on the heuristics used, while TTG gains are larger and less sensitive to model variations (c.f., Tables \ref{tab_ocval_r101}, \ref{tab_octest_r101}).

\begin{table}[t]
	\centering
		\caption{\small $\textbf{AP}_{mask}$ on COCOPersons \emph{val} (ResNet-101-FPN, 3 seeds). Same trends as ResNet-50-FPN, i.e., without a large domain-shift, TTA hurts and TTG gives a small gain ($1\%$).}
	\resizebox{0.3\textwidth}{!}{    
{
\begin{tabular}{|c|c|c|c|}
\hline
\multirow{2}{*}{\textbf{Model}}     & \multirow{2}{*}{\bfseries\makecell{Baseline}}     & \multirow{2}{*}{\bfseries\makecell{\textbf{TTA}}}     & \multirow{2}{*}{\bfseries\makecell{\textbf{TTG}}} \\
&   &   &   \\ \hline
Mask-RCNN     & \makecell{60.87\\(0.10)} & \makecell{59.22\\(0.21)} & \makecell{61.73\\(0.09)} \\ \hline
Variant 1     & \makecell{60.93\\(0.10)} & \makecell{59.14\\(0.10)} & \makecell{61.82\\(0.02)} \\ \hline
Variant 2     & \makecell{60.71\\(0.15)} & \makecell{59.49\\(0.06)} & \makecell{61.69\\(0.12)} \\ \hline
Variant 3     & \makecell{61.15\\(0.10)} & \makecell{59.70\\(0.28)} & \makecell{61.81\\(0.19)} \\ \hline
\end{tabular}}}
	\label{tab_coco_r101}
\end{table}

\begin{table}[t]
	\centering
		\caption{\small $\textbf{AP}_{mask}$ on OCHuman \emph{val} (ResNet-101-FPN, 3 seeds). Same trends as with ResNet-50-FPN, i.e., TTG gains are larger and less dependent on pose head's architectural nuances.}
	\resizebox{0.3\textwidth}{!}{    
{
\begin{tabular}{|c|c|c|c|}
\hline
\multirow{2}{*}{\textbf{Model}}     & \multirow{2}{*}{\bfseries\makecell{Baseline}}     & \multirow{2}{*}{\bfseries\makecell{\textbf{TTA}}}     & \multirow{2}{*}{\bfseries\makecell{\textbf{TTG}}} \\
&   &   &   \\ \hline
Mask-RCNN     & \makecell{19.21\\(0.36)} & \makecell{19.58\\(0.23)} & \makecell{22.67\\(0.35)} \\ \hline
Variant 1     & \makecell{19.10\\(0.40)} & \makecell{19.45\\(0.40)} & \makecell{22.60\\(0.23)} \\ \hline
Variant 2     & \makecell{19.32\\(0.38)} & \makecell{20.30\\(0.17)} & \makecell{22.51\\(0.32)} \\ \hline
Variant3      & \makecell{19.77\\(0.18)} & \makecell{20.46\\(0.25)} & \makecell{22.50\\(0.15)} \\ \hline
\end{tabular}}}
	\label{tab_ocval_r101}
\end{table}

\begin{table}[t]
	\centering
		\caption{\small $\textbf{AP}_{mask}$ on OCHuman \emph{test} (ResNet-101-FPN, 3 seeds). Same trends as with ResNet-50-FPN, i.e., TTG gains are larger and more consistent across various variants.}
	\resizebox{0.3\textwidth}{!}{    
{
\begin{tabular}{|c|c|c|c|}
\hline
\multirow{2}{*}{\textbf{Model}}     & \multirow{2}{*}{\bfseries\makecell{Baseline}}     & \multirow{2}{*}{\bfseries\makecell{\textbf{TTA}}}     & \multirow{2}{*}{\bfseries\makecell{\textbf{TTG}}} \\
&   &   &   \\ \hline
Mask-RCNN     & \makecell{18.76\\(0.02)} & \makecell{18.97\\(0.28)} & \makecell{21.95\\(0.54)} \\ \hline
Variant 1     & \makecell{18.71\\(0.27)} & \makecell{18.93\\(0.23)} & \makecell{22.42\\(0.22)} \\ \hline
Variant 2     & \makecell{18.92\\(0.19)} & \makecell{19.92\\(0.11)} & \makecell{22.09\\(0.58)} \\ \hline
Variant 3     & \makecell{19.68\\(0.05)} & \makecell{20.26\\(0.28)} & \makecell{22.27\\(0.38)} \\ \hline
\end{tabular}}}
	\label{tab_octest_r101}
\end{table}

Table \ref{tab_other_methods} compares the TTG $\textbf{AP}_{mask}$ for Variant 2 against some of the existing methods in the literature as described in Section \ref{sec:backgroundrelated} (numbers are cited from the referenced papers). While an apple to apple comparison is difficult due to the nuances in each paper setup, we observe that our simple TTG method provides a competitive performance especially noting that we report \emph{average} values across multiple seeds. This reconfirms our observation that TTG gains are less dependent on the heuristics used and more robust to architectural variations.

\begin{table}[t]
	\centering
		\caption{\small Comparison of TTG $\textbf{AP}_{mask}$ for Variant 2 against methods from the literature.}
	\resizebox{0.47\textwidth}{!}{    
{
\begin{tabular}{|c|c|c|c|c|}
\hline
\multirow{2}{*}{\textbf{Model}}     & \multirow{2}{*}{\textbf{Backbone}}     & \multirow{2}{*}{\bfseries\makecell{COCO \\Persons \emph{val}}}     & \multirow{2}{*}{\bfseries\makecell{OCHuman\\ \emph{val}}}     & \multirow{2}{*}{\bfseries\makecell{OCHuman\\ \emph{test}}} \\
&   &   &   &   \\ \hline
Mask-RCNN~\cite{he2017mask}          & ResNet-50-FPN  & \makecell{53.2}  & \makecell{16.3}     & \makecell{16.9} \\ \hline
Pose2Seg~\cite{zhang2019pose2seg}    & ResNet-50-FPN  & \makecell{55.5}  & \makecell{22.2}     & \makecell{23.8} \\ \hline
CondInst~\cite{condinst}             & ResNet-50-FPN  & \makecell{54.8}  & \makecell{20.3}     & \makecell{20.1} \\ \hline
YOLACT~\cite{bolya2019yolact}       & ResNet-101-FPN & \makecell{50.2}  & \makecell{13.2}     & \makecell{13.5} \\ \hline
Variant 2 (ours)                     & ResNet-50-FPN  & \makecell{60.8}  & \makecell{21.5}     & \makecell{20.9} \\ \hline
Variant 2 (ours)                     & ResNet-101-FPN & \makecell{61.7}  & \makecell{22.5}     & \makecell{22.1} \\ \hline
\end{tabular}}}
	\label{tab_other_methods}
\end{table}

\subsection{Ablation Studies}
\begin{figure}[t]
\begin{center}
\includegraphics[width=0.95\linewidth]{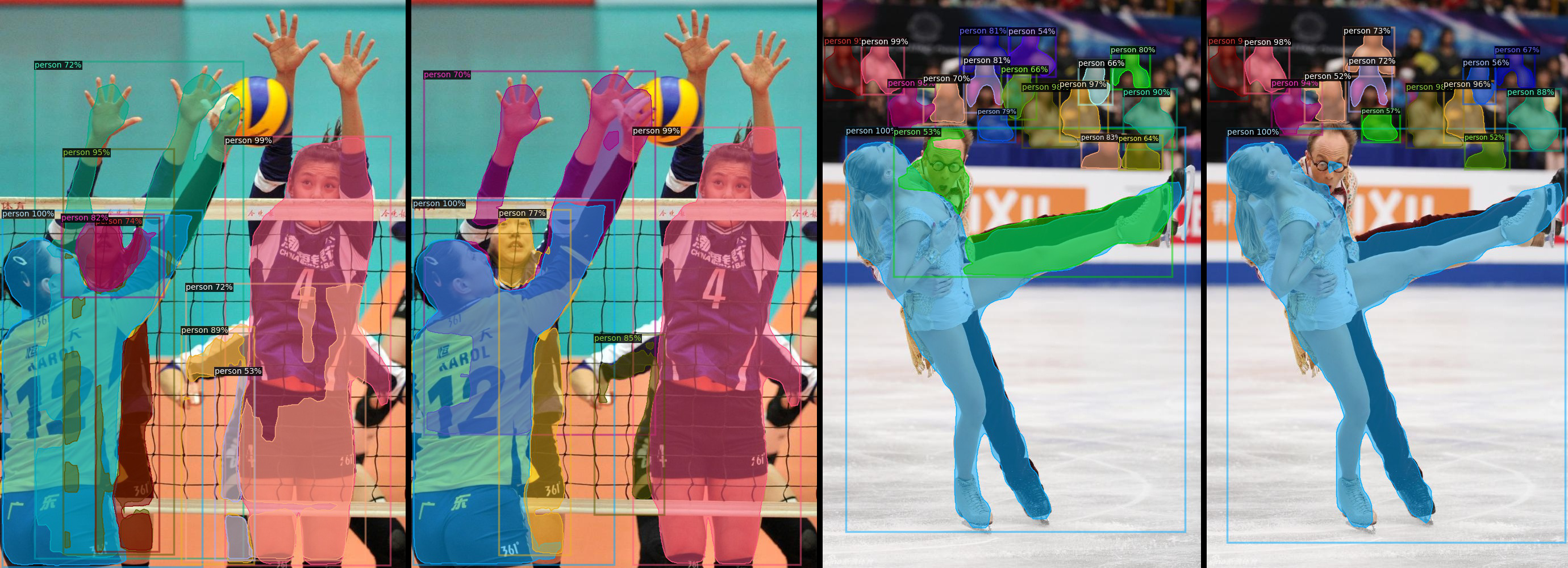}
\end{center}
   \caption{Increasing the min-person-score from $0.5$ to $0.8$ removes some false-positives from the volleyball scene (first and second from left, respectively), but also causes some true-positives to be missed in the ice-skating scene (third and fourth).}
\label{fig:min_person_score}
\end{figure}

In this section we report several ablation studies that shed light on some of the factors limiting the performance of test-time adaptation. Tables \ref{tab_ocval_r50_lr}, \ref{tab_ocval_r50_mps} and \ref{tab_ocval_r50_mkp} give the TTA $\textbf{AP}_{mask}$ for Variant 2 on OCHuman \emph{val} when sweeping the \emph{adaptation} learning-rate, min-person-score (i.e., threshold for rejecting all keypoints for a detected person) and min-keypoint-prob (i.e., threshold for rejecting an individual keypoint), respectively. As the tables show, our choices for these parameters (i.e., $1\mathrm{e}{-3}$, $0.5$ and $0.05$, respectively) lie within the relevant sweet-spots, however, since we do not assume any knowledge about the target domain, we cannot use the \emph{optimal} values (as Table \ref{tab_ocval_r50_lr} indicates a learning-rate of $1\mathrm{e}{-3}$ is not optimal) and have to rely on judgment calls which is an important limiting factor. For example as Figure \ref{fig:min_person_score} shows, increasing the min-person-score from $0.5$ to $0.8$ removes some of the false-positives from the volleyball scene, but also causes some true-positives to be missed in the ice-skating scene.

\begin{table}[t]
	\centering
		\caption{\small TTA $\textbf{AP}_{mask}$ on OCHuman \emph{val} with different learning rates (ResNet-50-FPN, 4 seeds).}
	\resizebox{0.3\textwidth}{!}{    
{
\begin{tabular}{|c|cccc|}
\hline
\multirow{2}{*}{\textbf{Model}} & \multicolumn{4}{c|}{\textbf{Lr}} \\ \cline{2-5} 
                                & \multicolumn{1}{c|}{\textbf{0.5e-3}} & \multicolumn{1}{c|}{\textbf{1e-3}} & \multicolumn{1}{c|}{\textbf{2e-3}} & \textbf{4e-3}\\ \hline
Variant 2                       & \multicolumn{1}{c|}{\makecell{18.49\\(0.29)} } & \multicolumn{1}{c|}{\makecell{18.75\\(0.29)} } & \multicolumn{1}{c|}{\makecell{18.92\\(0.21)} } & \makecell{18.57\\(0.38)} \\ \hline
\end{tabular}}}
	\label{tab_ocval_r50_lr}
\end{table}

\begin{table}[t]
	\centering
		\caption{\small TTA $\textbf{AP}_{mask}$ on OCHuman \emph{val} with different min-person-score (ResNet-50-FPN, 4 seeds).}
	\resizebox{0.3\textwidth}{!}{    
{
\begin{tabular}{|c|cccc|}
\hline
\multirow{2}{*}{\textbf{Model}}       & \multicolumn{4}{c|}{\textbf{Min. Person Score}} \\ \cline{2-5} 
                                      & \multicolumn{1}{c|}{\textbf{0.5}} & \multicolumn{1}{c|}{\textbf{0.6}} & \multicolumn{1}{c|}{\textbf{0.7}} & \textbf{0.8}\\ \hline
Variant 2                             & \multicolumn{1}{c|}{\makecell{18.78\\(0.25)} } & \multicolumn{1}{c|}{\makecell{18.70\\(0.27)} } & \multicolumn{1}{c|}{\makecell{18.74\\(0.26)} } & \makecell{18.70\\(0.31)} \\ \hline
\end{tabular}}}
	\label{tab_ocval_r50_mps}
\end{table}

\begin{table}[t]
	\centering
		\caption{\small TTA $\textbf{AP}_{mask}$ on OCHuman \emph{val} with different min-keypoint-prob (ResNet-50-FPN, 4 seeds).}
	\resizebox{0.3\textwidth}{!}{    
{
\begin{tabular}{|c|cccc|}
\hline
\multirow{2}{*}{\textbf{Model}} & \multicolumn{4}{c|}{\textbf{Min. Keypoint Prob.}} \\ \cline{2-5} 
                                & \multicolumn{1}{c|}{\textbf{0.05}} & \multicolumn{1}{c|}{\textbf{0.1}} & \multicolumn{1}{c|}{\textbf{0.2}} &\textbf{0.4} \\ \hline
Variant 2                             & \multicolumn{1}{c|}{\makecell{18.75\\(0.30)} } & \multicolumn{1}{c|}{\makecell{18.54\\(0.25)} } & \multicolumn{1}{c|}{\makecell{18.29\\(0.25)} } & \makecell{18.08\\(0.30)} \\ \hline
\end{tabular}}}
	\label{tab_ocval_r50_mkp}
\end{table}

\begin{figure}[t]
\begin{center}
\includegraphics[width=0.95\linewidth]{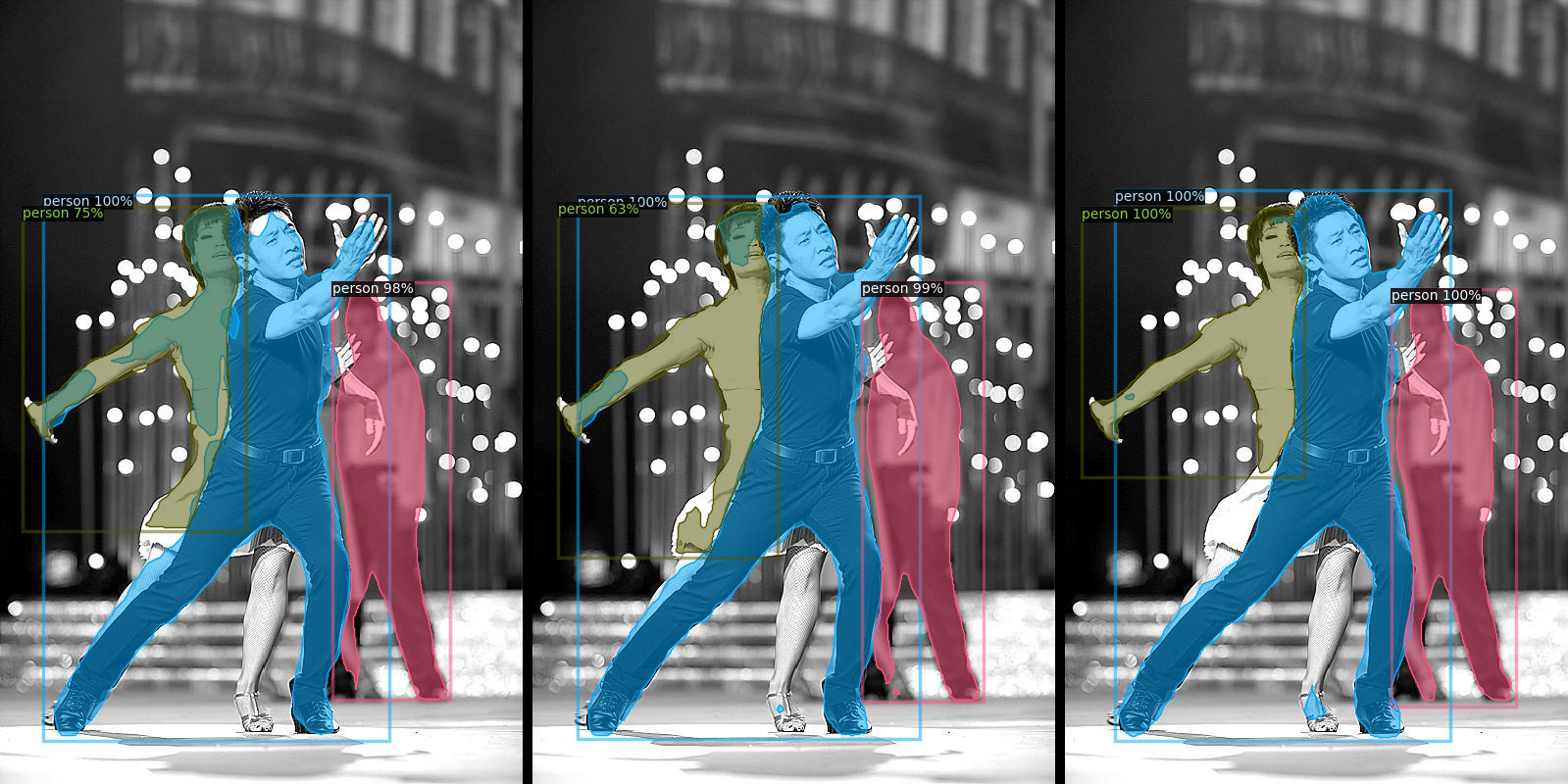}
\end{center}
   \caption{Dynamic selection of hyper-parameters, based on the test image, may improve TTA gain significantly, but is impractical in the absence of prior knowledge about the target domain. From left to right: baseline and test-time-adapted Mask-RCNN for TTA learning rates of $0.5\mathrm{e}{-3}$, and $4\mathrm{e}{-3}$, respectively. For this test image a learning rate of $0.5\mathrm{e}{-3}$ greatly cleans up the spill-over of the man's mask. A larger learning rate of $4\mathrm{e}{-3}$ degrades the performance by shrinking the woman's mask.}
\label{fig:learning_rate}
\end{figure}

We note that dynamic selection of hyper-parameters such as TTA learning-rate, depending on the test image at hand, may significantly improve the TTA performance, e.g., for the particular image shown in Figure \ref{fig:learning_rate}, a learning rate of $0.5\mathrm{e}{-3}$ greatly cleans up the spill-over of the man's mask. Whereas a larger learning rate of $4\mathrm{e}{-3}$ degrades the performance by shrinking the woman's mask. We also note that such dynamic hyper-parameter selection is impractical in the absence of prior knowledge about the target domain (i.e., another TTA limiting factor).  

Table \ref{tab_ocval_r50_lr_defrost} reports numbers for the scenario where the backbone is only partially adaptable (i.e., stage $5$ of ResNet-50-FPN). As the table shows, for smaller learning rates the partially adapted backbone yields smaller TTA gains compared to the fully-adaptable one. However for the large learning-rate of $4\mathrm{e}{-3}$ it retains more of the gain which is consistent with being partially adaptable, hence more regularized. 

\begin{table}[t]
	\centering
		\caption{\small TTA $\textbf{AP}_{mask}$ on OCHuman \emph{val} with backbone fully or partially adapted (ResNet-50-FPN, 4 seeds).}
	\resizebox{0.4\textwidth}{!}{    
{
\begin{tabular}{|c|cccc|}
\hline
\multirow{2}{*}{\textbf{Model}} & \multicolumn{4}{c|}{\textbf{Lr}} \\ \cline{2-5} 
                                & \multicolumn{1}{c|}{\textbf{0.5e-3}} & \multicolumn{1}{c|}{\textbf{1e-3}} & \multicolumn{1}{c|}{\textbf{2e-3}} & \textbf{4e-3}\\ \hline
\makecell{Variant 2 w. backbone \\
fully adapted}         & \multicolumn{1}{c|}{\makecell{18.49\\(0.29)} } & \multicolumn{1}{c|}{\makecell{18.75\\(0.29)} } & \multicolumn{1}{c|}{\makecell{18.92\\(0.21)} } & \makecell{18.57\\(0.38)} \\ \hline
\multicolumn{1}{|c|}{\makecell{Variant 2 w. backbone's \\
last-stage adapted}} & \multicolumn{1}{c|}{\makecell{18.34\\(0.37)} } & \multicolumn{1}{c|}{\makecell{18.57\\(0.39)} } & \multicolumn{1}{c|}{\makecell{18.85\\(0.28)} } & \multicolumn{1}{c|}{\makecell{18.83\\(0.35)} } \\ \hline
\end{tabular}}}
	\label{tab_ocval_r50_lr_defrost}
\end{table}

Finally, Table \ref{tab_ocval_r50_tta_ttg} gives $\textbf{AP}_{mask}$ numbers for the case where a TTG model is test-time adapted. We note that TTA gains, if any, are even smaller as TTG already takes advantage of the features generated by the keypoints head.

\begin{table}[t]
	\centering
		\caption{\small $\textbf{AP}_{mask}$ on OCHuman \emph{val} with both TTG and TTA (ResNet-50-FPN, 4 seeds). Applying TTA on top of TTG has minimal gain, if any, as TTG already uses keypoint features.}
	\resizebox{0.35\textwidth}{!}{    
{
\begin{tabular}{|c|c|c|c|}
\hline
\multirow{2}{*}{\textbf{Model}}     & \multirow{2}{*}{\bfseries\makecell{Baseline}}     & \multirow{2}{*}{\bfseries\makecell{\textbf{TTG}}}     & \multirow{2}{*}{\bfseries\makecell{\textbf{TTG+TTA}}} \\
&   &   &   \\ \hline
Mask-RCNN     & \makecell{17.74\\(0.25)} & \makecell{21.48\\(0.10)} & \makecell{21.37\\(0.22)} \\ \hline
Variant 1     & \makecell{17.93\\(0.19)} & \makecell{21.56\\(0.17)} & \makecell{21.77\\(0.27)} \\ \hline
Variant 2     & \makecell{18.15\\(0.27)} & \makecell{21.51\\(0.51)} & \makecell{21.70\\(0.41)} \\ \hline
Variant 3     & \makecell{18.41\\(0.29)} & \makecell{21.20\\(0.14)} & \makecell{20.96\\(0.43)} \\ \hline
\end{tabular}}}
	\label{tab_ocval_r50_tta_ttg}
\end{table}

\section{Conclusion}\label{sec:conclusion}
We compared two approaches for enhancing human segmentation masks using keypoints estimation. First was test-time adaptation (TTA), where we allowed test-time adjustments of network weights based on the unlabeled test image without access to the labeled source dataset. It worked by back-propagating keypoints estimates, as pseudo-labels, to adjust the backbone weights. The second approach was training-time generalization (TTG), where we allowed offline access to the labeled source dataset but no test-time alteration of weights. We further did not assume the availability of any test images. Our TTG method worked by augmenting the backbone features with that of the keypoints head and using it to infer masks. We also devised three additional pose-head variants to improve keypoint pseudo-label quality, using keypoint visibility/occlusion prediction, and use of global-attention (i.e., transformers).  

We evaluated both approaches and, through ablations, identified factors limiting the TTA gains, i.e., we showed that without a large domain shift, TTA hurts performance, and TTG shows only a small gain. In contrast, for a large domain shift, TTA gains are smaller and more heuristics dependent, while TTG's are larger and more robust.

{\small
\bibliographystyle{ieee_fullname}
\bibliography{egbib}
}

\end{document}